\documentclass[letterpaper, 10 pt, conference]{ieeeconf}  

\IEEEoverridecommandlockouts                              

\usepackage{graphicx} 
\usepackage{times} 
\usepackage{amsmath} 
\usepackage{amssymb}  
\usepackage{subfig}
\usepackage[ruled, linesnumbered]{algorithm2e}
\usepackage[draft]{hyperref}

\title{\LARGE \bf
Fast Footstep Planning on Uneven Terrain Using Deep Sequential Models
}

\author{\authorblockN{Hersh Sanghvi}
\authorblockA{School of Engineering and Applied Science\\
University of Pennsylvania\\
Email: hsanghvi@seas.upenn.edu}
\thanks{This work was supported by NSF grant CCF-2112665 (TILOS).}
\and
\authorblockN{Camillo Jose Taylor}
\authorblockA{School of Engineering and Applied Science\\
University of Pennsylvania\\
Email: cjtaylor@seas.upenn.edu}}

\begin{document}

\maketitle
\thispagestyle{empty}
\pagestyle{empty}

\begin{abstract}

One of the fundamental challenges in realizing the potential of legged robots is generating plans to traverse challenging terrains. Control actions must be carefully selected so the robot will not crash or slip. The high dimensionality of the joint space makes directly planning low-level actions from onboard perception difficult, and control stacks that do not consider the low-level mechanisms of the robot in planning are ill-suited to handle fine-grained obstacles. One method for dealing with this is selecting footstep locations based on terrain characteristics. However, incorporating robot dynamics into footstep planning requires significant computation, much more than in the quasi-static case. In this work, we present an LSTM-based planning framework that learns probability distributions over likely footstep locations using both terrain lookahead and the robot's dynamics, and leverages the LSTM's sequential nature to find footsteps in linear time. Our framework can also be used as a module to speed up sampling-based planners. We validate our approach on a simulated one-legged hopper over a variety of uneven terrains. 

\end{abstract}

\begin{figure}
    \centering
    \subfloat[]{\includegraphics[width=0.45\textwidth]{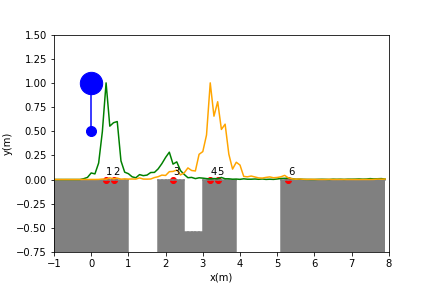}}
    \vspace{-1.0em}
    \subfloat[]{\includegraphics[width=0.45\textwidth]{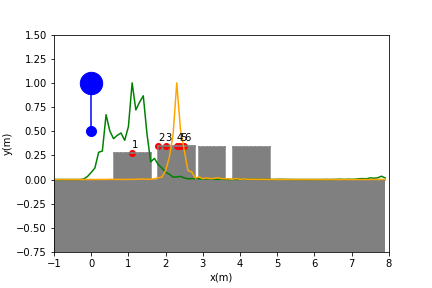}}\\
    \caption{Our footstep planner in action. The green and orange surfaces represent probability distributions over the terrain for the first and fourth steps, respectively. Note in particular how the planner accounts for the loss in kinetic energy on the elevated terrain.}
    \label{fig:distrs1}
\end{figure}

\section{Introduction}

Legged robots have significant mobility advantages over their wheeled counterparts. They can traverse stairs, uneven terrain and other hazards that would be impassable for more conventional ground robots. 
However, these attractive dynamic capabilities come at a cost. First, legged robots with limited perceptual lookahead must carefully plan motions to move across potentially rough terrain without crashing. This involves developing an effective mapping from perception to feasible actions that incorporates terrain geometry and the robot's high-dimensional and hybrid state dynamics. Second, factors such as model mismatch and perception error force frequent online replanning. These two requirements significantly constrain the possible approaches. Searching over the full state space would certainly result in a feasible plan, but would take far too long to be used online, while a planner using many approximations would operate quickly, but with riskier actions.


One commonly used approach is to plan footstep locations \cite{zucker_optimization_2011,kuindersma_optimization_2016,kalakrishnan_learning_2009}. Footstep planning frames the problem as a discrete-time search and also allows the planner to incorporate both the geometry of the terrain and the robot's dynamics while considerably lowering the dimensionality of the search space. Making this search fast enough to run online in dynamic situations is still a significant challenge. 

In this paper, we develop a framework that learns from offline plans to give explicit and interpretable probability distributions over the terrain conditioned both on perceptual input and the robot's dynamics for an arbitrary number of footsteps. This distribution can be used directly through its mode (argmax), or to speed up a sampling-based planner by suggesting informed samples. We demonstrate the merits of our approach using a simulated one-legged hopper on a variety of terrains. The proposed framework provides a computationally efficient approach to relating look-ahead perception to action for these systems.

\section{Related Work}

\subsection{Graph Search and Optimization for Footstep Planning}
Methods for footstep planning that do not use machine learning are typically based on solving either graph search or optimization problems. These methods often generate quasi-static trajectories or are too slow to be used online. On the graph search side, \cite{zeglin_control_1998} uses best-first search over input angles for a one-legged robot, sampling possible leg angles at each impact to find the next node. For multi-legged robots, \cite{arain_comparison_2013, winkler_planning_2015, kalakrishnan_fast_2010, wermelinger_navigation_2016} use a variety of graph search algorithms to plan only a Center of Mass (CoM) trajectory and use terrain heuristics to find footholds along that trajectory, while \cite{zucker_optimization_2011,hornung_anytime_2012, vernaza_search-based_2009} explicitly search over footsteps based on stable stance transitions. 

Approaches that use optimization often make use of some relaxation or approximation of the underlying dynamics to speed up computation time. \cite{winkler_gait_2018} express step lengths as gait phases to find center-of-mass and footstep locations using nonlinear optimization, at the cost of not enforcing some swing foot collisions. \cite{deits_footstep_2014} formulates footstep planning as a mixed-integer quadratic program and can solve for short footstep plans in under a second, but does not consider the robot dynamics. \cite{ponton_convex_2016} uses a convex relaxation of the dynamics in order to generate 2 to 4 step plans in roughly one second. \cite{kuindersma_optimization_2016, posa_direct_2013} consider more complete dynamic models but the whole body optimizations take minutes to solve. Most of these optimization schemes could benefit from an independent system for initializing the footstep locations which our method could provide.

\subsection{Learning for Footstep Planning}
Researchers have also utilized learning-based methods to plan footstep locations. One popular method is to use Reinforcement Learning (RL) and learn a policy from visual inputs to select the optimal next footstep or action \cite{tsounis_DeepGait_2020, peng_deep_2017, gangapurwala_rloc_2020, meduri_deepq_2020, peng_terrain_2016}. These methods can leverage the full robot model, but usually only plan footsteps or motion primitives over a short horizon, and sometimes rely on hand-selected terrain features in the reward function. Variational Autoencoders have also been explored for footstep planning to deal with multimodality: the fact that there are many feasible choices for steps, depending on the desired behavior \cite{melon_recedinghorizon_2021}.

Another approach is to incorporate learned modules into the more traditional planning frameworks. Previous methods include learning cost functions to speed up optimization or search \cite{deits_lvis_2019, yang_realtime_2021} learning forward dynamics models \cite{lin_efficient_2019}, adaptation of next footholds based on visual information and heuristics \cite{magana_fast_2019, kim_vision_2020}, and direct regression of footsteps \cite{melon_reliable_2020}.

Recurrent Neural Networks (RNNs) are an interesting architecture for this problem as they naturally handle the sequential, discrete-time nature of the problem. While RNNs have been used in policies for blind low-level control of legged robots \cite{siekmann_blind_2021} and in other planning domains \cite{kuo_deep_2018, bency_neural_2019}, to the authors' knowledge this is the first application of RNNs to the  footstep planning problem. Our framework has a number of advantages over other learned footstep planners: no terrain heuristics are used for training, the plan can be arbitrarily long up to the visual horizon, multimodality is handled by outputting a probability density map over all possible step locations, and it can be used as a sub-module in a traditional sampling planner.

\section{Problem Formulation and Approach} \label{formulation}
We envision a scenario where a legged robot has a limited perceptual lookahead and must plan its steps in order to cross the terrain in front of it without crashing. Specifying these step locations can provide enough information to generate a corresponding CoM trajectory \cite{zucker_optimization_2011}. The locations of the footsteps must be carefully chosen to be reachable while also avoiding collisions. 


First, we set up our problem by describing the dynamics of our simulated robots. Next, we establish two baseline search-based planners that 
search for footsteps over broken terrain while respecting the dynamics. We then show how we can use these planners to train a recurrent model to output similar plans in a fraction of the computational time. Lastly, we demonstrate the performance of our recurrent planner on a variety of terrains. 

\section{Modeling} \label{sec:models}
\subsection{Spring-Loaded Inverted Pendulum}
For our work, we consider the classic planar Spring Loaded Inverted Pendulum (SLIP) hopper \cite{schwind_spring_1998}: a one legged machine modeled as a massless, spring-loaded leg and foot attached to a point mass body. This model is used as a template \cite{full_templates_1999} to plan trajectories of higher degree of freedom robots such as bipeds \cite{apgar_fast_2018}.

The SLIP hopper is a hybrid system with two modes. When the hopper is in the \textit{flight} mode, the foot is not touching the ground and the only force experienced by the body is gravity. When the hopper is in the \textit{stance} mode, the foot is in contact with the ground and the body experiences gravity and a spring force acting through the leg. The instantaneous transition from stance to flight is referred to as \textit{liftoff}, and the transition from flight to stance is \textit{touchdown}.

\begin{figure}
    \centering
    \includegraphics[width=0.9\linewidth]{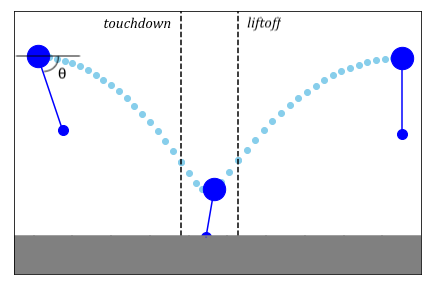}
    \caption{A typical SLIP hopper trajectory}
    \label{fig:slip_traj}
\end{figure}


The passive SLIP model has one degree of freedom: the angle of the leg with respect to the ground. This input is applied at the apex of flight, where the vertical velocity is 0, and will affect the body's trajectory during the following stance phase by controlling the tradeoff between apex height and forward velocity. The touchdown transition occurs when the foot makes contact with the ground. A prototypical trajectory for the SLIP hopper is shown in Figure \ref{fig:slip_traj}. For the full dynamical equations, we refer the reader to \cite{carver_lateral_2009}, whose formulation we used for this work.


An important property of the SLIP model is that there is no closed form solution for the stance phase dynamics, requiring either an approximation or explicit forward integration in order to calculate the next state after applying a control input \cite{piovan_enforced_2012}. 

\subsection{SLIP Failure Cases} \label{sec:failures}
Although the passive SLIP models are relatively simplistic, there are a number of failure cases that make planning for the SLIP hopper nontrivial over broken terrain. The first is the standard kinematic case where the body collides with the terrain, or the foot collides with a vertical surface. 
Second, the leg angle can exit the friction cone during stance phase, causing the foot to slip. Third, if the hopper's apex height is too high above the terrain, the spring will fully compress during touchdown and will bottom-out, also causing a crash. Finally, because the energy of the passive SLIP model is fixed, there is an inverse relationship between the apex velocity and apex height. If the hopper travels too fast, the apex height will not be sufficient for the foot to swing to its position for the next step, causing the robot to "trip" on the ground.


The intersection of the kinematically feasible and dynamically reachable sets of footsteps is not immediately apparent just by analyzing the terrain. Therefore, when planning a trajectory, one must either precompute a set of feasible motion primitives for varying terrains, perform search by sweeping over control inputs and simulating the dynamics, or solve a nonlinear optimization problem. 

\section{Baseline Planners} 
\subsection{Angle-space Planner}
We formulate our first baseline planner as a best-first search over a tree that is dynamically constructed by sampling the possible actions at each leaf, adapted from the approach in \cite{zeglin_control_1998}. Each node $N$ is a tuple $(a, x, \theta, P, c, \{C_1...C_N\})$, where $a$ is the apex state at which the leg angle $\theta$ is applied to reach $x$, the footprint of the CoM at the start of the subsequent stance phase. The other terms contain information about the graph structure. $P$ is the parent of $N$, $c$ is the cost of $N$, and $\{C_1...C_N\}$ is the list of the children of $N$.

Our search algorithm begins with a root node, which is the inevitable touchdown point from the initial apex. From that node, we sample all the available actions and simulate the dynamics equations. If an action does not result in one of the failure cases outlined in Section \ref{sec:failures}, a node corresponding to the current apex state and subsequent touchdown location is added to the tree as a leaf. Afterwards, the node with the lowest cost is fetched from a priority queue and expanded. This process repeats until a desired number of nodes exceed the goal distance. We refer to this planner as the "Angle-space" planner. Pseudocode for this algorithm is shown in Algorithm \ref{algo:search1}.


\begin{algorithm}
\SetAlgoLined
\SetKwInOut{Inputs}{Inputs}
\SetKwInOut{Output}{Output}

\Inputs{Initial apex state $a$, terrain model $t$, goal $g$, cost function $h$, forward model $f$, branching factor $B$, desired number of sequences $d$}
\Output{Up to $d$ sequences of feasible touchdown locations}
 Set $N_{\text{root}} = (a,\, x_0)$\;
 Initialize empty priority queue $q$\;
 $N_s$ $\gets$ $N_{root}$\;
 \While{\# successful solns $<$ $d$ \textit{and} q not empty}{
  $a_s = N_s.a$\;
  Generate $\{\theta_1, ..., \theta_B\}$ within friction limits of $t$\;
  \For{$i = 1$ \KwTo $B$} {
    \uIf{successful(\,f$(a_s, \,t, \,\theta_i$))} {
        $a_{s+1}, x_{s+1} \gets f(a_s, \,t, \,\theta_i)$\;
        $c \gets h(x_{s+1})$\;
        $N_{s+1} \gets (a_{s+1}, x_{s+1}, \theta_i, N_s, c, \{\})$\;
        Add $N_{s+1}$ as a child of $N_s$\;
        Add $N_{s+1}$ to $q$\;
    }
  }
  $N_s \gets q.$pop()\;
  \uIf{atGoal($N_s, \,g$)} {
    \# successful solns++\;
    Add $N_s$ to goal list\;
    $N_s \gets q.$pop()\;
  }
 }
 \For{$N$ in goal list} {
    Output all $x$ on the path from $N_{root}$ to $N$\;
 }
 \caption{Angle-space Planner}
 \label{algo:search1}
\end{algorithm}


To evaluate the cost of each transition, we use a function that takes the child node as input. This cost function is formulated as 
\begin{align}
    h(N_i) = w_1|x_i - x'| + w_2|\dot{x}_i| + w_3D(N_{i-1}, N_{i}, N_{i+1})
\end{align}
Where $x_i$ is the node $N_i$'s location and $x'$ is the goal location. This cost function penalizes the distance of the node to the goal, high node apex velocity, and the node's "isolation". $D(N_{i-1}, N_i, N_{i+1})$ is a function calculating the spread between the nearest neighbors of $N_i$. If the $N_i$ obtained by applying $\theta_i$ is far away from from its closest fellow child, that means a small deviation from $\theta_i$ is more likely to cause a crash in the presence of obstacles. $(w_1, w_2, w_3)$ are the relative weights of these terms.

\subsection{Low-level Controller}
While the planner shown in Algorithm \ref{algo:search1} also produces the input angles necessary to hit the footstep sequence it outputs, it is inadvisable to use these in practice, since the open-loop angles will be incorrect as soon as there is any deviation from the reference sequence. 
Our approach is similar in spirit to \cite{yim_precision_2018}; we train a simple Multi-Layer Perceptron (MLP) that is called once per apex. The training data for the MLP is generated by gridding over the space of current velocity $\dot{x}$, height $z$, and leg angle $\theta$, and running a simulation to obtain the resulting step length $x_L$. Then, $\dot{x},\, z,\, \text{and}\, x_L$ are fed as inputs to the MLP and the output is the required leg angle $\theta$.


Naturally, the learned controller does not perfectly achieve the desired step lengths. It can produce errors of up to 0.5m on our simulated robot, since at high velocities small angle errors can cause large position errors. 


\subsection{Step-space Planner}
One fault of the planner shown in Algorithm \ref{algo:search1} is that it has no knowledge of the underlying controller and will output sequences that are physically achievable by the robot, but may cause significant errors if provided as inputs to the actual system consisting of an imperfect controller composed with the dynamics. 

To incorporate the controller model, we alter the sampling procedure used in line 6 of Algorithm \ref{algo:search1}. Instead of directly generating leg angles, we instead generate a grid of possible step lengths ${x_L}$, pass those with the current apex state into the controller to obtain the desired $\theta$, and use it in the planning algorithm. The branching factor for this variant corresponds to the size and spacing of the grid ${x_L}$. The final sequence output by this modified planner uses the sampled step lengths $x_L$, as those feedforward inputs are known to result in feasible actions when input to the controller. This approach has the advantage that steps that encounter terrain obstacles can be avoided a priori, but requires a model for the low-level controller which is not always known in practice. We refer to this modification as the "Step-space planner". 



\section{Learned Recurrent Planner}

\begin{figure}
    \centering
    \includegraphics[width=0.9\linewidth]{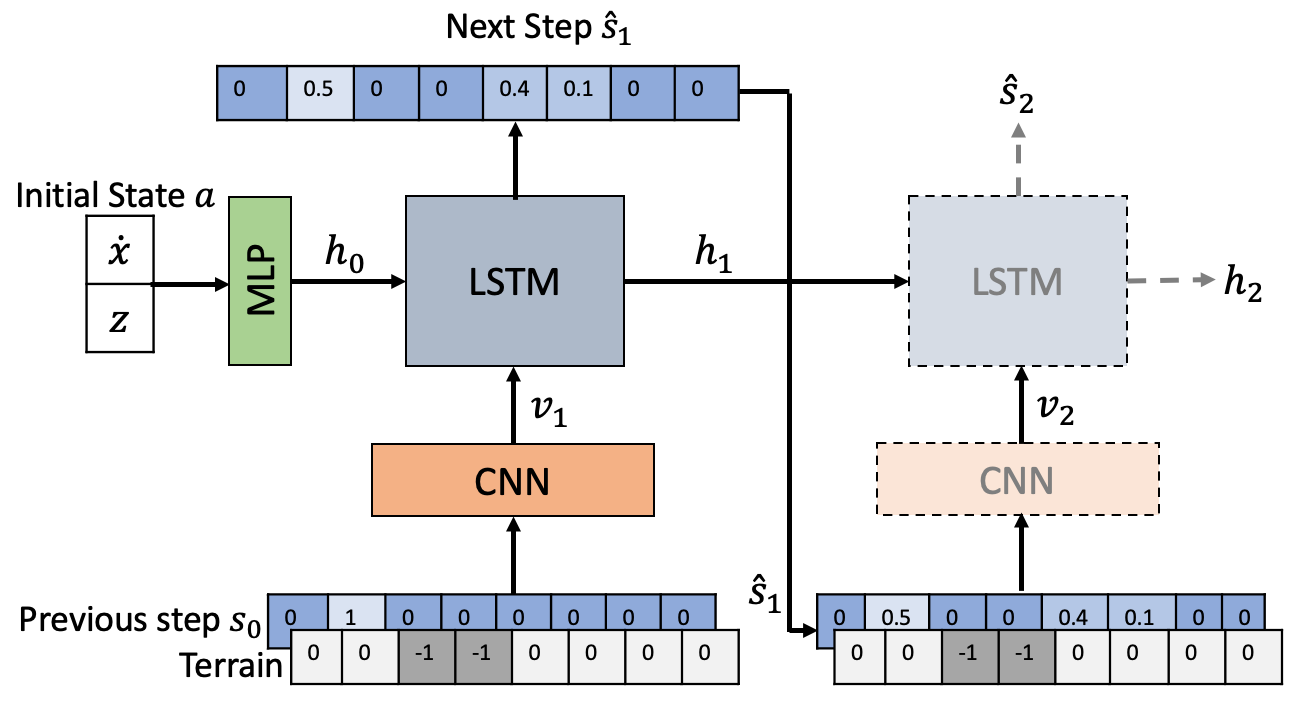}
    \caption{A diagram of our LSTM framework unrolled over time}
    \label{fig:arch}
\end{figure}

Both of the search planners described in the previous section operate too slowly to be used online with frequent replanning. For a fast, online footstep planner, we instead use a learned sequential framework to generate plans that mimic the ones output by the search planners. 

We use a Long Short-term Memory network \cite{hochreiter_long_1997} (LSTM), a variant of RNNs, as the core of our learned planner, with a Convolutional Neural Network (CNN) that jointly processes the terrain and previous step location to create an encoded vector as input to the LSTM.



Just like in sentence completion (for which RNNs are widely used), a large history of footstep sequences over a wide variety of terrains and initial apexes induces a probability distribution over possible next step locations: $p(s_{i} | t, a, s_0,...,s_{i-1})$, 
where $t$ is a discretized encoding of the perceived terrain, $a$ is the initial apex prior to step $s_0$, and $s_0,...,s_{i-1}$ are the past steps. This distribution can be used for planning either by sampling next nodes from the distribution as in \cite{kuo_deep_2018, qureshi_deeply_2018}, or by taking the mode of the distribution as the next footfall location.

All together, our full LSTM-CNN model is governed by the following equations:
\begin{align*}
    h_0 &= g_0(a) \\
    v_i &= \text{CNN}([\hat{t}, \hat{s}_{i-1}]) \\
    \hat{s}_i, h_i &= \text{LSTM}(h_{i-1}, v_i)
\end{align*}
Where $h_0$ is the initial apex $a$ projected onto the hidden state space of the LSTM by $g_0$, a single linear layer. $v_i$ is the encoded input vector, calculated by the CNN by processing the previous step $\hat{s}_{i-1}$ and a discretized array representation of the terrain $\hat{t}$. In a typical call to the LSTM planner, $s_0$ is the inevitable touchdown location at the current apex, and it contains a "1" at some location and "0" everywhere else because its location is fixed. However, the future steps $\hat{s}_{1:n}$ used for recurrent planning are uncertain, so they are nontrivial discrete probability densities. For a cell in the heightmap $\hat{t}$ with index $j$, $\hat{s}_i[j]$ is the probability that the $i$-th step lands on $j$. A diagram of our model is shown in Figure \ref{fig:arch}.

Also, since the steps $\hat{s}$ share the same underlying support as the heightmap $\hat{t}$, finding the correct step location becomes a classification problem among the discrete cells. By encoding the step location in the terrain, the CNN can use co-located visual features to process both together as channels of a single image. 



The LSTM architecture allows the learned planner to capture the behavior of the hopper over time without requiring simulation of the forward dynamics, which is a key to handling the complex dynamics of the system without excessive computation. Meanwhile the CNN allows the planner to extract relevant terrain features and to marry that with an aligned representation of the planned footsteps.




\subsection{LSTM-Guided Sampling}
We can also modify the Step-space planner to sample next footsteps from the output probability distribution of the LSTM instead of from a fixed grid. To generate these samples, we fetch the path from the root node to the current node, and pass the path into the LSTM as the past steps, along with the initial apex and terrain. 

Sampling from the output of the LSTM allows us to reduce the required branching factor of the Step-space planner by sampling from an informed distribution, while also incorporating a model of the controller into the LSTM planner. We use temperature scaling \cite{guo_calibration_2017} to widen the output distribution and make it more suitable for sampling.

\section{Experiments}
We test our approach on the robot described in Section \ref{sec:models}. For all experiments, we use a hopper whose body mass is 7kg, leg length is 0.5m, and spring constant is 3200 $N m^{-1}$. For simulation, we use a custom simulation of the SLIP model with a timestep of 0.01 seconds. All of the sequence generation, model training, experiments, and simulations were done in Python on a machine with an AMD Ryzen 9 3600 processor with 32 GB of RAM and an NVidia RTX 2080 Super GPU.

Our LSTM model contains 2 layers each of 110 hidden units, and the input CNNs use 2 layers of 7x1 filters before being encoded into a 110-dimensional feature vector that acts as input to the LSTM, making our overall architecture relatively small. Because we frame planning as a classification problem, we use Cross-Entropy Loss to train our model.

\subsection{Training}
We generate training data for our LSTM by running the Angle-space planner with the parameters described in Table \ref{table:planar_training_params}. A single run of the Angle-space planner can give multiple training sequences as any path from the root node to a leaf is a feasible sequence of footsteps. In order to still provide reasonably good training sequences, we only add sequences that make a certain amount of progress towards the goal to the training set. We use the Angle-space planner instead of the Step-space planner to prevent the LSTM from overfitting to the performance of any particular low-level controller.

Each terrain is either a ditches or steps scenario. We model the terrain 8 meters in front of the robot and 3 meters behind, discretized in segments of 10cm. We also use a lower coefficient of friction during training in order to encourage further robustness by the LSTM. This process generates roughly 50,000 sequences of varying lengths (although a few of the sequences on the same terrain are usually highly similar to each other) and takes 1.5 hours with multithreading. 

\subsection{Testing}
Our testing data consists of 15 random ditch-world and step-world terrains coupled with 5 different initial states swept over varying initial velocities (150 total test cases). Given that the train and test sets use completely random terrains and initial states, it is highly unlikely that any of the test terrains appear in the training set. The coefficient of friction used for all the test experiments is 0.8. Similar to training, the goal for the robot is to cross a certain $x$ threshold, set to 10 meters.


\begin{table}[]
\centering
\caption{Training dataset generation parameters}
\begin{tabular}{|l|l|}
\hline
\textbf{Parameter}                 & \textbf{Value} \\ \hline
\# different terrains               & 360   \\ \hline
\# initial states per terrain     & 8     \\ \hline
Min \# steps       & 8     \\ \hline
\# desired goal sequences per instance  & 8     \\ \hline
Branching Factor          & 15    \\ \hline
Coefficient of friction   & 0.6   \\ \hline
$w_0$, $w_1$, $w_2$       & 1, 1, 2 \\ \hline
\end{tabular}
\label{table:planar_training_params}
\end{table}


We test 5 different planners: the Angle-space planner, Step-space planner, LSTM planner, LSTM-Guided sampling planner, and a terrain-heuristic planner. The terrain-heuristic planner is meant to mimic the approaches of \cite{magana_fast_2019, kim_vision_2020}, where steps are selected based on a nominal gait dependent on the initial apex velocity, and then the planned steps are moved if they land in ditches. All planners are tested in a receding-horizon fashion, where the planned sequence is executed by the low-level controller for 3 steps before replanning occurs. 

\section{Results}
\begin{figure*}[h]
    \centering
    \subfloat[]{\includegraphics[width=\textwidth]{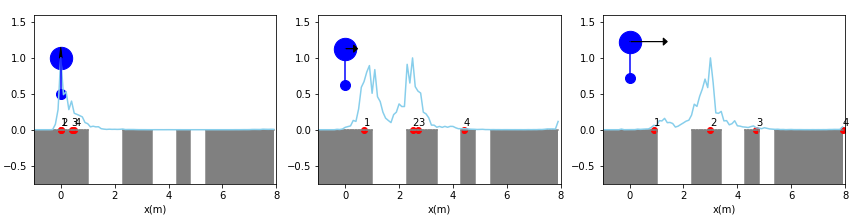} \label{fig:qualditches}}
    \vspace{-1.0em}
    \subfloat[]{\includegraphics[width=\textwidth]{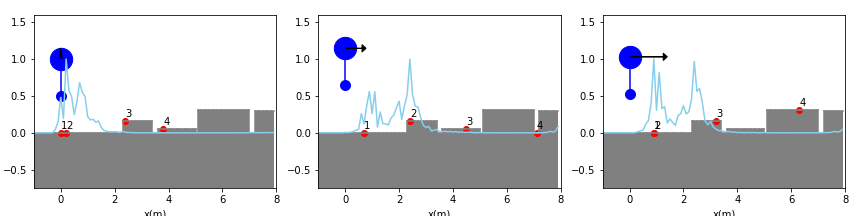} \label{fig:qualsteps}}
    \caption{Probability distributions over possible \textbf{second} step locations predicted by the LSTM. Increasingly long arrows indicate higher initial energy. Distributions are plotted such that the mode has height 1.0.}
    \label{fig:qualitative}
\end{figure*}

We measure success rate, computation time, and the number of calls made to the simulator (the ODE solver) by the Angle-space planner, LSTM planner, Step-space planner, LSTM-Guided planner, and a heuristic baseline planner on ditch and step test suites. We also discuss some qualitative behaviors of our planners and cases where they can be improved.
\subsection{Quantitative Results}

\begin{table}[]
\caption{Results on the Ditch Test Suite}
\begin{tabular}{|l|lll|}
\hline
\textbf{Planner}     & \textbf{Success $\%$} $\uparrow$ & \textbf{ODE Calls} $\downarrow$ & \textbf{$t_{\text{plan}}$} $\downarrow$ \\ \hline
Heuristic   & 24         & \textbf{0}              & \textbf{2e-5s}               \\ \hline
Angle-space & 58.6       & 22661          & 0.15s               \\ \hline
Step-space  & \textbf{86.6}       & 30857          & 0.29s               \\ \hline
LSTM (Ours)       & 56         & \textbf{0}              & 0.007s              \\ \hline
LSTM-Guided (Ours) & 84       & 4885           & 0.04s               \\ \hline
\end{tabular}
\label{table:ditches}
\vspace{1.0em}
\caption{Results on the Step Test Suite}
\begin{tabular}{|l|lll|}
\hline
\textbf{Planner}     & \textbf{Success $\%$} $\uparrow$ & \textbf{ODE Calls} $\downarrow$ & \textbf{$t_{\text{plan}}$} $\downarrow$ \\ \hline
Heuristic   & 56         & \textbf{0}              & \textbf{2e-5s}               \\ \hline
Angle-space & 60       & 27021          & 0.15s               \\ \hline
Step-space  & 85       & 41146          & 0.34s               \\ \hline
LSTM (Ours)       & 49        & \textbf{0}              & 0.007s              \\ \hline
LSTM-Guided (Ours) & \textbf{96}       & 6932            & 0.04s               \\ \hline
\end{tabular}
\label{table:steps}
\end{table}

The success rate of all 5 planners on the ditch and step world scenarios are shown in Tables \ref{table:ditches} and \ref{table:steps}. In these results, the Angle-space planner is used with an initial branching factor of 20 with a fallback branching factor of 30 if the first try fails. The Step-space planner is used with a horizon of 5m and spacing of 0.5 meters (branching factor of 10), and the LSTM-Guided planner is used with a branching factor of 3 and fallback factor of 5. 

We can see that the Step-space and LSTM-Guided planners have the best overall performance, succeeding on over 80\% of the terrains from both suites. This makes intuitive sense as they incorporate the controller's behavior into the planner. The pure LSTM planner performs slightly worse than the baseline Angle-space planner in both the ditches and steps scenarios; at low velocities, it tends to suggest steps that either cause the controller to violate the friction cone or that are very close to the edges of ditches and steps. However, the high performance of the LSTM-Guided sampler shows that the LSTM still suggests a high-quality probability distribution over possible locations. The heuristic planner sometimes performs well but is highly sensitive to variations in its parameters.

Tables \ref{table:ditches} and \ref{table:steps} show that the LSTM-Guided planner achieves high success rates while making fewer than 1/6 the number of ODE calls than the Step-space planner, and fewer than 1/4 of the Angle-space planner. Since it guides sampling to a few areas, the LSTM-Guided planner has to explore fewer nodes to find a feasible plan, and so has a planning time of 0.04s on average, compared to 0.3s for the Step-space planner.



\subsection{Qualitative Analysis}
Figure \ref{fig:qualitative} shows 4-step plans and the probability distributions for the second step generated by the LSTM over 3 different initial apex states. 
As we increase both the apex height and initial velocity, the LSTM alters its plan accordingly to traverse the terrain. On the ditch terrain, the peaks of the distribution for the second step move further away as the initial energy increases. On the step terrain, when starting with too high an initial velocity it shortens its first two steps due to the tradeoff mentioned in Section \ref{sec:failures}. In the 3rd subfigure in Figure \ref{fig:qualsteps}, with too high an initial velocity, it will not be able to generate the necessary height to clear the obstacles, so it takes a short step to slow down.

The LSTM planner does have a number of shortcomings. 
Although it takes noticeably shorter steps when the hopper has lower energy, it still tends to propose more aggressive plans than the Angle-space planner. 
In addition, many of the pathological behaviors present in the training data also appear in the learned sequences, including stepping very close to edges, as seen in Figure \ref{fig:qualsteps}. In Figure \ref{fig:qualditches}, there are spikes in the probability distribution around the edges of the islands, corresponding to training sequences where the Angle-space planner selects the node that minimizes the distance to the goal. This occurs in part because we chose not to manually bias the planner using terrain heuristics.

\section{Conclusion and Future Work}

In this work, we demonstrated how we can use an LSTM model to learn a distribution for likely footsteps conditioned both on terrain geometry and robot dynamics, and that this model can be successfully used to quickly generate footstep plans and to speed up a sampling-based planner. 

The most immediate extension of this work is to scale to multi-legged robots in 3D. We have explored adapting our approach to a 3D SLIP model with 2D heightmaps, and have promising, yet preliminary results with careful adjustments to the loss function and network structure. Since sampling-based planners scale poorly to higher-dimensional systems, we are also experimenting with using accurate but slow trajectory optimizers for bipeds to generate long trajectories for our method. Future work remains in understanding how much data is required to capture the range of behaviors that more complicated robots are capable of, and implementing our method on real hardware.

\bibliographystyle{IEEEtran}
\bibliography{IEEEabrv,references}

\end{document}